# A Knowledge-Informed Large Language Model Framework for U.S. Nuclear Power Plant Shutdown Initiating Event Classification for Probabilistic Risk Assessment

**Min Xian[1], Tao Wang[1], Sai Zhang[2], Fei Xu[2], Zhegang Ma[2*]**
[1]University of Idaho
1776 Science Center Drive, Idaho Falls, ID 83402, USA
mxian@uidaho.edu, wang0080@vandals.uidaho.edu
[2]Idaho National Laboratory
2525 Fremont Ave., MS 3850, Idaho Falls, ID 83415-3850, USA
sai.zhang@inl.gov, fei.xu@inl.gov, zhegang.ma@inl.gov

*Corresponding author: zhegang.ma@inl.gov

# A Knowledge-Informed Large Language Model Framework for U.S. Nuclear Power Plant Shutdown Initiating Event Classification for Probabilistic Risk Assessment

**Abstract**

Identifying and classifying shutdown initiating events (SDIEs) is critical for developing low power shutdown probabilistic risk assessment for nuclear power plants. Existing computational approaches cannot achieve satisfactory performance due to the challenges of unavailable large, labeled datasets, imbalanced event types, and label noise. To address these challenges, we propose a hybrid pipeline that integrates a knowledge-informed machine learning mode to prescreen non-SDIEs and a large language model (LLM) to classify SDIEs into four types. In the prescreening stage, we proposed a set of 44 SDIE text patterns that consist of the most salient keywords and phrases from six SDIE types. Text vectorization based on the SDIE patterns generates feature vectors that are highly separable by using a simple binary classifier. The second stage builds Bidirectional Encoder Representations from Transformers (BERT)-based LLM, which learns generic English language representations from self-supervised pretraining on a large dataset and adapts to SDIE classification by fine-tuning it on an SDIE dataset. The proposed approaches are evaluated on a dataset with 10,928 events using precision, recall ratio, $F_1$ score, and average accuracy. The results demonstrate that the prescreening stage can exclude more than 97% non-SDIEs, and the LLM achieves an average accuracy of 93.4% for SDIE classification.

# 1. INTRODUCTION

A shutdown initiating event (SDIE) in a nuclear power plant (NPP) probabilistic risk assessment (PRA) is an event during shutdown that challenges plant control and safety systems whose failure could potentially lead to core damage. SDIEs are critical and require successful mitigation because they may further cause safety system failures, operator errors, core damage, and radioactive release [1,2]. Identifying and classifying SDIEs is essential for safety and risk assessment.

Conventionally, analysts manually conduct SDIE identification through reviewing the nuclear operating experience (OpE) data such as the component failure event reports in the Institute of Nuclear Power Operations (INPO) database for U.S. NPPs or the licensee event reports (LERs) that an NPP must submit to the Nuclear Regulatory Commission (NRC) if any events listed in the Part 50.73 of Title 10, Code of Federal Regulations (10 CFR 50.73) [3] occurred in the plant. While suitable for processing a small, manageable number of event reports, this manual reviewing approach and process could be time-consuming, expensive, and prone to human errors when we aim to identify and classify SDIEs from a large number of historical event reports. Two semi-supervises approaches [4] were proposed to reduce the false positives of anomaly detection using data sparsely labeled by using the condition reports. The approaches were evaluated using two simulated datasets. Recent works [5–11] have explored using machine learning (ML) and natural language processing (NLP) approaches for automated event analysis, which uses simulated sensor signals for fault diagnostic. The long-short-term memory (LSTM) [12] and gated recurrent unit (GRU) [13] networks were applied, and nine initiating events were identified from 823 simulated data samples [5]. A deep-learning-based approach [6] is proposed for event identification and signal reconstruction and simulated 395 initiating events from 12 event categories. A spatiotemporal feature extractor was trained to efficiently identify 12 classes of NPP initiating events [7] from a set of 125 simulated events.

A k-nearest neighbors' approach [8] is proposed to classify events into 13 event types. A hybrid framework [14] was proposed to extract event causality from LERs; the authors built a text corpus with 20,129 samples from 92 LERs, built a deep-learning-based approach for causal relation detection, and developed a knowledge-based approach to extract explicit cause-effect pairs. A comprehensive review [10] explored the applicability of artificial intelligence (AI) and ML techniques in various fields of the nuclear industry, such as reactor system design and analysis, plant operation and maintenance, and nuclear safety and risk analysis.

However, developing AI/ML models to identify SDIEs from failure event reports is challenging. First, failure event reports were prepared in different formats with very different levels of detail, and the contents were organized differently. Second, the existing SDIE datasets are small (e.g., only about 200 samples), which are insufficient to train pure data-driven models that learn patterns and knowledge from texts. Third, the SDIE dataset used in this work is extremely imbalanced, and SDIEs only account for 1.75% of all failure events. The dominant non-SDIEs will bias conventional AI/ML models trained using the dataset. Fourth, some event types of the existing labeled SDIEs might be characterized incorrectly, which could mislead the model training. Recent work [15] demonstrated good average performance by using text vectorization and a support vector machine classifier. It initialized research to create an NLP pipeline that could classify SDIEs automatically and illustrated the promising potential to improve conventional manual procedures. However, the results also showed that it was difficult to correctly classify events from categories with a small number of training samples.

To address the challenges, we propose a two-stage framework that uses a knowledge-informed large language model (LLM) to improve the performance of SDIE classification. The proposed approach consists of two stages: non-SDIE prescreening and SDIE classification. The prescreening stage detects possible SDIEs by using SDIE pattern-based text features. The SDIE patterns consist of the most important patterns of six SDIE types (ISOL, FLOW, LOCA,

LOAC, LOOP, and SFP; see **Table 1** for the definitions of these different SDIE types) and are prepared and grouped by domain experts. The second stage builds a Bidirectional Encoder Representations from Transformers (BERT)-based language model for SDIE classification. The model is pretrained using a large English language dataset and learns generic language patterns in a self-supervised manner. The pretrained BERT model is further fine-tuned using the SDIE dataset in a supervised fashion to classify input events into four types: ISOL&FLOW, LOOP, LOAC, and non-SDIE by combining ISOL and FLOW (which sometimes are hard to distinguish from each other) and removing SFP and LOCA (which have very few event counts).

The rest of this paper is organized as follows: Section 2 presents the components of the proposed framework and applies them to SDIE detection and classification; Section 3 demonstrates and discusses the experimental results of SDIE prescreening and classification; Section 4 discusses the challenges and possible strategies for further improving the SDIE classification OpE data analysis; and Section 5 concludes the paper along with the main contributions of the work.

**Table 1**. Event definitions and counts.

| Event Name | Description | Details | Event Count |
|---|---|---|---|
| **ISOL** | Trip or Isolation of Shutdown Cooling Loop | Primary isolation, does not include low-level trip due to LOCA | 27 |
| **FLOW** | Diversion or Loss of Cooling Water Flow | Blockage or diversion of primary coolant or service/closed cooling water flow path such that heat removal is no longer accomplished, does not include primary isolations or losses of primary coolant from the primary system | 23 |
| **LOCA** | Loss of Coolant Accident | Includes inadvertent drain-down of primary system where sufficient coolant no longer is available for the normal decay heat removal process | 13 |
| **LOAC** | Loss of Safety or Vital Bus for SDC Equipment | Loss of vital bus due to LOOP or local fault | 89 |
| **LOOP** | Loss of Offsite Power | Loss of electrical power to all unit safety buses requiring all emergency power generators to start and supply power to the safety buses. | 54 |
| **SFP** | Loss of Spent Fuel Pool Cooling | Loss of spent fuel pool cooling | 6 |
| **Non-SDIE** | Non-shutdown IE | Not related to shutdown initiating event | 10,716 |
| **Total** | | | 10,928 |

## 2. THE PROPOSED METHOD

### 2.1 Knowledge-Informed LLM Framework for Nuclear OpE Data Analysis

In this study, a novel framework with two processing stages is proposed to enable the application of LLM tools to tasks with extremely imbalanced text datasets, e.g., nuclear OpE data. The first stage builds a knowledge-based text vectorization using a task-related vocabulary and excludes non-task-related data by training a binary classifier. Domain experts create the task-related vocabulary to captures the most salient task-related patterns, which boosts the binary classifier's ability to exclude non-task-related data and create a more balanced dataset. The second stage leverages LLM which was pretrained using large datasets of English texts to understand generic language patterns. The LLM can be further fine-tuned to learn task-related knowledge and patterns, e.g., SDIE classification. The overall framework in the work can be used to explore the LLMs' capabilities in other nuclear OpE data analysis and NLP tasks, e.g., identifying and characterizing at-power initiating events or component failure events from the plant event reports, or exploring the digital instrumentation and control (DI&C) system-related software or hardware failures for DI&C reliability analysis.

We apply the proposed framework to identify and classify SDIEs from a large dataset of event descriptions. In the first stage, a vocabulary of 44 SDIE patterns is prepared and applied to extract the most representative feature vectors from event descriptions. A binary classifier is trained using the feature vectors to identify SDIEs and exclude non-SDIEs. Because of the first stage, a dataset with more balanced SDIEs and non-SDIEs is created. A pretrained LLM is fine-tuned on the dataset to classify the event types.

**2.2 Dataset and Preprocessing**

The dataset used in this work is from the INPO component failure database and the NRC Integrated Data Collection and Coding System (IDCCS) shutdown initiating event database that is based on the LERs, ranging from 12/9/1991 to 11/27/2021. It has **212** samples of SDIEs and **10,716** non-SDIEs, and each sample contains the event description and the event type. Analysts manually identified and classified all existing SDIE event types and labeled them using a new web-based NLP tool developed by the team. The tool enables collaborations among multiple users to manage projects, annotate SDIE types, input notes, and export results. **Table 1** shows the number of SDIE events from the IDCCS database for different SDIE types.

The original event text contains characters and string patterns for formatting, e.g., '\n,' '_0x00D_,' '***,' multiple whitespaces, and empty lines. While these formatting markers enable user-friendly content, they are a distraction for the computer algorithms developed to understand the text. A text cleaning pipeline is implemented to clean the data, and it consists of removing format characters and strings, whitespace, stop words, and lemmatization.

**2.3 SDIE Pattern-based Text Vectorization**

The patterns of SDIEs are a set of keywords and phrases that are commonly used to describe SDIEs and are used as task-specific vocabulary to convert texts to quantitative feature vectors. If a large set of SDIE samples was provided, an automatic process could be developed to identify the vocabulary by just using the frequencies of words. However, in our dataset, SDIEs account for less than 2% of all samples, which is insufficient to develop an automatic process to build a good vocabulary. Furthermore, in this specialized domain, the vocabulary is closely related to the facilities and designs of NPPs. Therefore, two senior nuclear researchers manually built and refined a set of SDIE patterns.

**Table 2**. SDIE patterns. The patterns are organized into seven categories. Phrases or words in "()" are sub-patterns.

| Event Name | SDIE Patterns | Pattern Notation |
|---|---|---|
| SD mode | Mode 3, Mode 4, Mode 5, Mode 6, No Mode, Cold Shutdown, Hot Shutdown, (Refueling Outage, Refuel Outage), (Defuel, Defueled) | $P_0$-$P_8$ |
| Loss of SDC | (loss of shutdown cooling, loss of SDC), (loss of RHR, loss of Residual Heat Removal), (loss of decay heat removal, decay heat removal was lost), (shutdown cooling, Shutdown Cooling, SDC), (Residual Heat Removal, RHR), decay heat removal | $P_9$-$P_{14}$ |
| LOAC | loss of AC, partial loss of offsite power, loss of voltage, (Emergency Diesel Generator, EDG), (Engineered Safety Features, ESF), (emergency bus, vital bus, essential bus, safeguard bus, safety bus, 4160v bus, 4.16kv bus), (Alternating Current, Alternate Current, AC), (de-energized, de-energizing, deenergized, deenergizing) | $P_{15}$-$P_{22}$ |
| ISOL and FLOW | Primary Containment Isolation, containment isolation, (isolation of shutdown cooling, isolation of Shutdown Cooling, isolation of SDC), isolation valve, RHR pump, running RHR, operating RHR, running Residual Heat Removal, operating Residual Heat Removal), (isolated, isolation), trip, (closure, closed), (actuation, actuated), (interrupted, interruption) | $P_{23}$-$P_{33}$ |
| LOCA | (LOCA, Loss of Coolant), (draining, draindown, inadvertent draindown), reactor cavity, water level, Reactor Coolant System, RCS), spray pump | $P_{34}$-$P_{39}$ |
| LOOP | (LOOP, loss of offsite power, loss of off-site power), (loss of power, power loss), (loss of 230 kv, loss of 230kv) | $P_{40}$-$P_{42}$ |
| SFP | (Spent Fuel Pool, spent fuel cooling, SFP) | $P_{43}$ |

**Table 2** shows the text patterns defined for SDIEs to extract more event-related features from raw text. The patterns include a set of 44 keywords and phrases ($P_0$-$P_{43}$) and could reduce the impact of non-relevant texts and models' dependency on large datasets. The quantization process searches the patterns in each group, records the number of occurrences, and generates a feature vector, $\boldsymbol{x} = (x_0, x_1, \cdots, x_{43})^\mathrm{T}$, for each data sample. $x_k$ is defined by

$$x_k = \sum_{j=1}^{n} f(t, P_{k,j}) \qquad (1)$$

where $t$ denotes a text sample, $P_{k,j}$ is the $j$th sub-pattern of $P_k$, and $f(t, P_{k,j})$ is the pattern frequency of $P_{k,j}$ in text sample $t$.

## 2.4 Non-SDIE Prescreening

In the first stage, a logistic regression model is trained to classify input feature vectors into two categories: non-SDIEs and possible SDIEs. Since the dimension of the feature vector is only 44, logistic regression is sufficient to achieve satisfactory results. A weighted binary cross-entropy loss with the L² norm regularizer is defined as the training objective

$$L(w) = \sum_{i=1}^{n}[-y_i \cdot logp_i - (1-y_i)log(1-p_i)] + \alpha \cdot \|w\|_2^2 \qquad (2)$$

$$p_i = \frac{1}{1 + e^{(-w^T x_i + b)}} \qquad (3)$$

where $w = (w_0, w_1, \cdots, w_{43})^T$ denotes the vector of model parameters, b is the bias, $y_i \in \{0, 1\}$ is the ground truth of the $i$th event, $p_i$ is the prediction, and hyperparameter $\alpha$ defines the trade-off between the cross-entropy loss and the L² norm.

## 2.5 LLM-based SDIE Classification

In the second stage, we developed a BERT-based LLM [16] for classifying SDIE types. BERT is a transformer-based [17] model pretrained on a large corpus of English datasets. It was pretrained in a self-supervised manner to predict masked words in sentences (i.e., masked language modeling) and predict if two masked sentences follow each other or not. The BERT model works as a foundation LLM that learns the inner representation of the English language and can be applied to different downstream tasks. By applying the pretrained BERT model that has learned generic language patterns, we could significantly reduce the required number of annotated samples from our task.

The pretrained BERT model outputs a vector of 768 features, and we concatenate the model with a dropout layer (30% dropout rate) and an output layer with four units. The final model has about 110 million parameters. The cross-entropy loss and the Adam optimizer are used to fine-tune the model on our dataset for SDIE classification.

## 3 EXPERIMENTAL RESULTS

### 3.1 Experiment Setup and Evaluation Metrics

In the prescreening, the weighted binary cross-entropy/log loss with $\mathbf{L^2}$ norm is applied as the loss function, and the logistic regression model is trained using the stochastic gradient decent method. The hyperparameter $\boldsymbol{\alpha}$ is set to 1e-4 by experiments. The class weight for event type non-SDIE and SDIE is set to 0.019 and 0.981, respectively. The classifier is trained using 70% of the samples of the whole dataset and tested using 30% of the samples.

In the SDIE classification, the proposed LLM uses the cross-entropy loss and Adam optimizer with a learning rate of 1e-5. The LLM is trained using an Nvidia V100 GPU. Since the SDIE dataset is small, the 5-fold cross-validation is applied to exploit the whole dataset fully and produce a reliable evaluation of model performance. The dataset is divided into five disjoint subsets of the same size. Each subset serves as a test set while the others are used for training in rotating order, i.e., five models are trained, and the final model performance is calculated by accumulating test results from the five models. This entire training process is repeated for 50 epochs with early stopping.

The precision, recall ratio, $\mathbf{F_1}$ score, and average accuracy are used to evaluate the performance of the proposed approaches. They are defined by

$$\mathbf{precision} = \frac{\mathbf{TP}}{\mathbf{TP} + \mathbf{FP}} \qquad (4)$$

$$\mathbf{recall} = \frac{\mathbf{TP}}{\mathbf{TP} + \mathbf{FN}} \qquad (5)$$

$$\mathbf{F_1} = \frac{2 \cdot \mathbf{precision} \cdot \mathbf{recall}}{\mathbf{precision} + \mathbf{recall}} \qquad (6)$$

$$\mathbf{accuracy} = \frac{\mathbf{TP} + \mathbf{TN}}{\mathbf{n}} \qquad (7)$$

where TP is the number of true positives in the predicted results of an event type, TN is the number of true positives, FP is the number of false positives, FN is the number of false negatives, and n denotes the size of the dataset (e.g., training set or test set). The precision, recall ratio, and $F_1$ score are calculated for each class or event type; and the accuracy is calculated on a whole dataset.

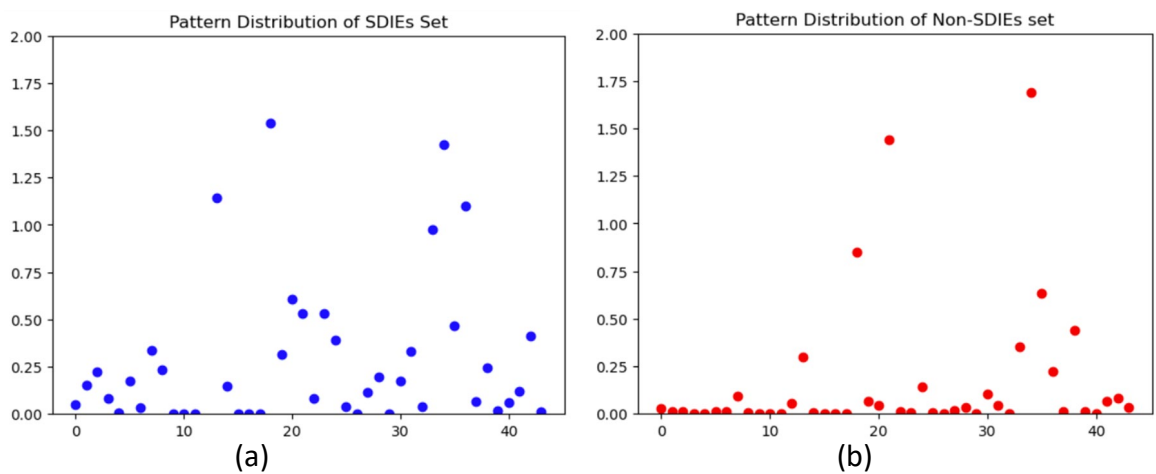

**Figure 1**. The distributions of 44 SDIE patterns on the (a) SDIEs and (b) non-SDIEs sets. The value of each pattern is calculated using the average number of occurrences of the pattern on the dataset. The horizontal axis is the indices of SDIE patterns. See **Table 2** for the details of SDIE patterns.

### 3.2 SDIE Prescreening

In the prescreening, we apply handcrafted SDIE patterns to extract quantitative features (i.e., number of occurrences of each pattern) that can distinguish SDIEs from non-SDIEs. As shown in **Figure 1**, the pattern distributions on the two datasets have significant differences, e.g., $P_1$, $P_2$, $P_5$, $P_7$, $P_8$, $P_{12}$, $P_{13}$, $P_{19-24}$, and $P_{42}$ of SDIEs are larger than those in non-SDIEs, which indicate that the most features from the two groups are highly separable.

The prescreening stage trains a logistic regression model to classify input event into SDIE or non-SDIE category. It achieves an average accuracy of 94.7%, and 95.9% on the training and test set, respectively. Notably, on the test set, 96.1% (3,094/3,221), non-SDIEs are identified correctly, and only seven SDIEs are misclassified. Note that the precision and $F_1$ score of the SDIE category are low on both the training set, but the metrics are biased by the imbalance nature the dataset, i.e., SDIE samples only account for less than 2% of the whole dataset. For example, on the test set, even though only 3.9% (127) of non-SDIEs are misclassified into SDIEs, the small number of all SDIEs (58) produces low precision and $F_1$ values. The results demonstrate that the proposed features are highly effective in distinguishing SDIEs from non-SDIEs; and the high recall values of the non-SDIE category show that the proposed model could exclude more than 97% of non-SDIEs.

### 3.3 LLM-based SDIE Classification

Table 3. Results of the prescreening stage.

|  |  | SDIE | Non-SDIE | Total |
|---|---|---|---|---|
| **Training set** | Metric | 154 | 7,495 | 7,649 |
|  | predicted* | 138/525 | 7,108/7,124 | 7,246/7,649 |
|  | precision | 26.3% | 99.8% | – |
|  | recall | 89.6% | 96.0% | – |
|  | $F_1$ | 40.7% | 98.0% | – |
|  | accuracy | – | – | 94.7% |
| **Test set** | Metric | 58 | 3,221 | 3,279 |
|  | predicted* | 51/178 | 3,094/3,101 | 3,146/3,279 |
|  | precision | 28.7% | 99.8% | – |
|  | recall | 87.9% | 96.1% | – |
|  | $F_1$ | 43.3% | 97.9% | – |
|  | accuracy | – | – | 95.9% |

*A/B: A refers to the number of correctly detected samples; B refers to the total number of predicted samples (i.e., TP + FP) in a category.

After the prescreening stage, the 5-fold cross-validation is used to validate the performance of the proposed method reliably on a small dataset. The dataset consists of 193 SDIEs and 314 non-SDIEs. In the original 212 SDIE samples, the SFP (6) and the LOCA (13) event types are removed because LLM cannot learn meaningful insights from the small number of data samples. Also, the ISOL and FLOW types are combined into one category (i.e., ISOL&FLOW) because the two types are closely related and even nuclear experts cannot reach a consensus yet on distinguishing the two event types. The 314 non-SDIEs are predicted as possible SDIEs from the prescreening stage and are input into the LLM in the second stage. The preprocessing step reduces the number of non-SDIEs by more than 95%, which significantly mitigates the imbalanced issue between SDIEs and non-SDIEs. For example, the percentage of SDIEs has increased from 1.93% to 34.52% in the refined dataset.

As shown in **Table 4**, the proposed LLM achieves an overall accuracy of 93.4% in classifying four types of events on the dataset. It obtains outstanding performance in recognizing event types ISOL&FLOW, LOAC, and non-SDIE. Even though 314 non-SDIEs are classified as suspicious SDIEs in the prescreening stage, all of them are correctly recognized as non-SDIE in this stage. The performance of the LOOP type is reasonably good but not as outstanding as the other types. It is caused by significant text overlap between LOOP and LOAC, which leads to similar word embeddings and increases the chance of misclassifying them. The high overall accuracy and $F_1$ scores demonstrate the effectiveness of the proposed LLM.

Table 4. Results of SDIE classification using 5-fold cross-validation.

|  | ISOL&FLOW | LOAC | LOOP | Non-SDIE | Total |
|---|---|---|---|---|---|
| **# of Events** | 50 | 89 | 54 | 314 | 507 |
| **predicted*** | 45/50 | 76/90 | 43/54 | 314/314 | 476/507 |
| **recall** | 90.0% | 85.4% | 79.6% | 100% | – |
| **precision** | 90.0% | 84.4% | 79.6% | 100% | – |
| $F_1$ | 0.90 | 0.85 | 0.80 | 1.0 | – |
| **accuracy** | – | – | – | – | 93.4% |

*A/B: A refers to the number of correctly detected samples; B refers to the total number of predicted samples (i.e., TP + FP) in a category.

## 4 DISCUSSION

This work achieves promising results and demonstrates that a hybrid, knowledge-informed LLM could address the significant challenges in the SDIE classification. The following discussion offers perspectives for further improving the SDIE classification in the future.

LLMs have demonstrated extraordinary capability to learn complex linguistic patterns and structures in text data in many applications, such as chatbots, medical support, coding, and writing. The nuclear field has accumulated massive historical operating experience data, and future nuclear-specific LLMs could be developed to explore the rich information and improve fine-grained SDIE classification, e.g., generate both primary and secondary (if available) SDIEs and build the causal chain of events.

One significant challenge faced during model development is that the number of text samples from some categories (e.g., SFP and LOCA) is small. The optimization processing during the model training tends to misclassify these categories. We attempted to ease this issue by applying a weighted loss function that gives more penalties for the misclassification of the small categories. In the long run, the ultimate solution will be to collect and/or generate more samples of these categories.

We observed that some events were likely labeled inaccurately, e.g., some of the ISOL events could be mislabeled as FLOW events. These data samples could confuse the training of ML models and could lead to imperfect performance. Further efforts are needed to create a large dataset with accurate event labels.

## 5 CONCLUSION

In this work, we propose a knowledge-informed LLM framework for SDIE detection and classification. The proposed approach integrates an SDIE patterns-based prescreening stage and an LLM-based SDIE classification stage and achieves outstanding performance on both non-SDIE detection and SDIE classification. The main contributions are (1) building a set of 44 SDIE patterns and a text vectorization approach, (2) developing an accurate non-SDIE prescreening approach that can significantly reduce the number of non-SDIEs, and (3) proposing an LLM method that classifies four event types accurately.